%% file: main.tex
\documentclass[sigconf]{acmart}

\graphicspath{{figures/}}
\usepackage{enumitem}
\usepackage{mwe}
\usepackage{array}
\usepackage{bbm}
\usepackage{amsmath}
\usepackage{algorithm}
\usepackage{algpseudocode}
\usepackage{pgfplots,pgfplotstable}
\usepackage{booktabs}
\usepackage{balance}
\usepackage{multirow}
\usepackage{subcaption}
\pgfplotsset{compat=newest}
\usepackage{fancybox}
\usepackage[most]{tcolorbox}
\usepackage{lipsum}
\usepackage{graphicx}
\usepackage{cleveref}
\usepackage{float}
\usepackage{pifont}
\usepackage{tikz}
\usetikzlibrary{arrows.meta,positioning,fit,backgrounds,calc}
\usetikzlibrary{shapes.geometric}
\usepackage[table]{xcolor}
\usepackage{xspace}

\newcommand{\sysname}{GraphMind\xspace}

\newcommand{\system}{GraphMind\xspace}
\newcommand{\company}{Microsoft\xspace}

\setcopyright{none}
\renewcommand\footnotetextcopyrightpermission[1]{}
\begin{document}
\title{\system: From Operational Traces to Self-Evolving Workflow Automation}

\author{Yiwen Zhu}
\affiliation{\institution{Microsoft, USA}}
\email{yiwzh@microsoft.com}

\author{Joyce Cahoon}
\affiliation{\institution{Microsoft, USA}}
\email{jcahoon@microsoft.com}

\author{Anna Pavlenko}
\affiliation{\institution{Microsoft, USA}}
\email{anna.pavlenko@microsoft.com}

\author{Qiushi Bai}
\affiliation{\institution{Microsoft, USA}}
\email{qiushibai@microsoft.com}

\author{Nima Shahbazi}
\authornote{Work done during an internship at Microsoft.}
\affiliation{\institution{University of Illinois Chicago, USA}}
\email{nshahb3@uic.edu}

\author{Divya Vermareddy}
\affiliation{\institution{Microsoft, USA}}
\email{dvermareddy@microsoft.com}

\author{Meina Wang}
\affiliation{\institution{Microsoft, USA}}
\email{meiwa@microsoft.com}

\author{Mathieu Demarne}
\affiliation{\institution{Microsoft, USA}}
\email{mdemarne@microsoft.com}

\author{Swati Bararia}
\affiliation{\institution{Microsoft, USA}}
\email{barariaswati@microsoft.com}

\author{Wenjing Wang}
\affiliation{\institution{Microsoft, USA}}
\email{wenjwang@microsoft.com}

\author{Hemkesh Vijaya Kumar}
\affiliation{\institution{Microsoft, USA}}
\email{hvijayakumar@microsoft.com}

\author{Hannah Lerner}
\affiliation{\institution{Microsoft, USA}}
\email{hannahlerner@microsoft.com}

\author{Katherine Lin}
\affiliation{\institution{Microsoft, USA}}
\email{katlin@microsoft.com}

\author{Steve Toscano}
\affiliation{\institution{Microsoft, USA}}
\email{stoscano@microsoft.com}

\author{Miso Cilimdzic}
\affiliation{\institution{Microsoft, USA}}
\email{misoc@microsoft.com}

\author{Subru Krishnan}
\affiliation{\institution{Microsoft, Spain}}
\email{subru@microsoft.com}

\begin{abstract}
Complex operational workflows coordinating personnel, tools, and information are central to system operations, yet end-to-end automation remains challenging due to extensive human input requirements and limited ability to adapt over time.
We present \sysname, a system that constructs, executes, and evolves action-centric workflow graphs with minimal human effort. The system operates in three phases. First, a \emph{scalable offline pipeline} extracts structured workflow graphs from large volumes of human resolution traces, capturing problems, actions, and their causal relationships. Second, an \emph{online multi-agent traversal} engine navigates the graph to dynamically construct and execute workflows, combining graph-guided retrieval with LLM-driven reasoning at each step. Third, \emph{Adaptive Traversal Reinforcement} (ATR) reinforces successful traversal paths, enabling execution-informed graph adaptation.
\sysname has been deployed across four production cloud database services for incident investigation. Evaluated on 93 held-out incidents and validated via blind expert review, the system outperforms an Agentic Summary-RAG baseline in mitigation reach, hallucination rate, and diagnostic throughput while requiring 8$\times$ less retrieval context. The ATR layer reduces hallucination rate by 26\%, demonstrating that workflow graphs can learn from execution feedback. A 12-week field study confirms practical value: 97\% of scored conversations yield actionable results within interactive latency.
\end{abstract}

\maketitle
\renewcommand{\shortauthors}{Yiwen Zhu et al.}

\input{1-introduction}
\input{2-related_works}
\input{3-offline_construction}

\input{4-online}
\input{4a-aco_evolution}

\input{5-evaluation}
\input{5a-offline_eval}
\input{5b-online_eval}
\input{5c-aco_eval_v2}
\input{6-deployment}
\input{6a-discussion}
\input{8-conclusions}
\input{9-genai-disclosure}

\bibliographystyle{ACM-Reference-Format}
\bibliography{sample}

\end{document}

%% file: 1-introduction.tex
\section{Introduction}
Organizations rely on complex operational workflows to coordinate personnel, tools, and information across domains like customer support and service operations.
While linear tasks are easily scripted, end-to-end automation remains challenging due to knowledge-intensive processes spanning heterogeneous tools and requiring iterative, context-dependent decisions. For example, diagnosing a database failure involves querying telemetry, analyzing timelines, drilling into exceptions, and ultimately applying mitigations---each step informed by prior observations.
Although LLM-based agents~\cite{brown2020language, openai2023gpt4, yao2023react, wang2024survey, schick2023toolformer} and agent skills (e.g., markdown-based instructions in Claude Code~\cite{claudecode} and GitHub Copilot CLI~\cite{githubcopilotcli}) have renewed interest in workflow automation, purely prompt-driven approaches remain brittle: manually authored knowledge artifacts are costly, difficult to maintain, inefficient at retrieval time, and unable to capture the long tail of edge cases. Meanwhile, operational traces---records of how experts actually carried out workflows---remain underutilized as structured knowledge.

In this work, we demonstrate that \textbf{action-centric workflow graphs}, constructed from operational traces, offer a far more efficient and agent-friendly format. Moreover, these graphs can learn from their own execution via \textbf{Adaptive Traversal Reinforcement} (ATR), inspired by \textbf{Ant Colony Optimization (ACO)}~\cite{dorigo1996aco, dorigo2004aco}: 
successful paths are reinforced while stale edges decay.
As Table~\ref{tab:approach_comparison} summarizes, this design uniquely combines storage efficiency, agent-friendly traversal, zero human authoring, expandable coverage, and self-evolution.

\begin{table}[t]
\centering
\caption{Knowledge formats for workflow automation.}
\vspace{-0.3cm}
\label{tab:approach_comparison}
\setlength{\tabcolsep}{1.9pt}
\tiny
\begin{tabular}{@{}l*{4}{>{\centering\arraybackslash}p{1.4cm}}@{}}
\toprule
\textbf{Aspect} & \textbf{Playbooks-RAG} & \textbf{Agent Skills} & \textbf{Trace-RAG$^{\dagger}$} & \textbf{\sysname} \\
\midrule
Storage efficiency & \ding{55}\xspace Redundant & \ding{55}\xspace Per-skill & \ding{55}\xspace Per-trace & \ding{51}\xspace Shared nodes \\
Agent friendliness & \ding{55}\xspace Verbose/Noisy & $\sim$ Structured & \ding{55}\xspace Verbose/Noisy & \ding{51}\xspace Graph traversal \\
Human authoring & High & High & None & None \\
Coverage & Fixed top-$k$ & Fixed per-skill & Fixed top-$k$ & Grows with traces \\
Self-evolution & \ding{55} & \ding{55} & \ding{55} & \ding{51}\xspace Via ATR \\
\bottomrule
\multicolumn{5}{@{}l}{\scriptsize $^{\dagger}$Trace-RAG retrieves raw traces via similarity search; raw traces are verbose and noisy.}
\end{tabular}
\vspace{-0.5cm}
\end{table}

\subsubsection*{\textbf{Introduction to \sysname}}
\begin{figure}[t]
    \centering
    \includegraphics[width=\columnwidth]{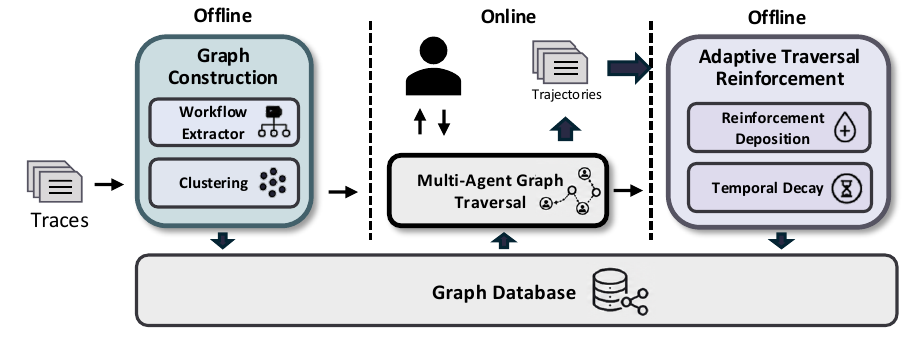}
    \vspace{-0.7cm}
    \caption{Three-phase architecture of \sysname.}
    \label{fig:architecture_intro}
    \vspace{-0.4cm}
\end{figure}

We present \sysname, a system realizing this vision through three pillars (Figure~\ref{fig:architecture_intro}).

\textit{Pillar 1: Offline graph construction.}
The offline pipeline extracts \textbf{actions} as the central unit of a lightweight \textbf{workflow graph} with a fixed schema, occupying a middle ground between raw-trace ingestion (high online cost) and entity-centric knowledge graphs~\cite{edge2024graphrag, han2025graphrag} (high offline cost and construction complexity), reducing online retrieval to bounded graph traversal (Section~\ref{sec:offline}). An \textbf{open-source} implementation is available at \url{https://aka.ms/graphmind}.

\textit{Pillar 2: Online multi-agent traversal.}
The online phase performs \emph{iterative graph traversal}: the agent retrieves a relevant subgraph, reasons over context and history, and selects the most promising action (\textbf{exploit}) or pivots to a different graph region (\textbf{explore}). Each traversal produces a \emph{trajectory} for reinforcement (Section~\ref{sec:online}).

\textit{Pillar 3: Adaptive Traversal Reinforcement (ATR).}
Inspired by ACO, we treat trajectories as reinforcement signals. \emph{Deposition} strengthens edges along successful paths; \emph{temporal decay} fades stale elements. The graph adapts over time through execution-informed reinforcement, encoding collective decision-making experience without manual engineering (Section~\ref{sec:aco}).

\subsubsection*{\textbf{Application to incident management.}}
We instantiate \sysname for incident management in large-scale cloud database services at \company, deployed in production across four services.

\subsubsection*{\textbf{Contributions}}
We present \sysname, to our knowledge the first production-deployed system that jointly constructs, executes, and reinforces action-centric workflow graphs from operational traces, requiring minimal manual authoring. Deployed across four production cloud database services, \sysname outperforms an Agentic Summary-RAG baseline that shares the same iterative agent orchestration in mitigation reach, hallucination rate, and expert-rated quality while requiring 8$\times$ less retrieval context (mean 3.9K vs.\ 32K tokens per graph load), demonstrating that workflow graphs can continuously learn and improve from autonomous agent execution at scale.

%% file: 2-related_works.tex
\section{Related Work}

\smallskip\noindent\textbf{ACO and swarm intelligence.}
ACO was introduced for combinatorial optimization~\cite{dorigo1996aco, dorigo2004aco} and extended in variants such as ACS and MAX-MIN~\cite{dorigo1997acs, stutzle2000maxmin}, with applications to routing~\cite{dicaro1998antnet}, scheduling~\cite{merkle2002aco_scheduling}, and feature selection~\cite{tabakhi2014aco_feature}. These methods optimize over fixed, well-defined search spaces with clear objective functions; \sysname adapts the reinforcement-and-decay intuition to evolving workflow graphs where the topology itself changes through online execution.

\smallskip\noindent\textbf{Process mining and knowledge graphs.}
Classical process mining recovers workflows from event logs~\cite{vanderaalst2004workflow,vanderaalst2012processmining} but assumes structured, machine-parseable inputs; \sysname targets noisier natural-language traces using LLM extraction. Prior work combines LLMs with knowledge graphs through fine-tuning or retrieval~\cite{pan2024unifying,kgsft2025}, while GraphRAG~\cite{han2025graphrag,lewis2020rag,ograg2024} methods use graph structure to improve retrieval. KG refinement via link prediction~\cite{bordes2013transe,dettmers2018conve,wang2014transr} learns static representations from fixed training data; LLMs can generate structured queries for grounded context~\cite{kovriguina2023sparql,shahkgqa} but do not update the underlying graph.

\smallskip\noindent\textbf{AIOps and LLM agents.}
D-Bot, Panda, and RCACopilot apply LLMs to troubleshooting~\cite{zhou2024dbot,singh2024panda,chen2024rcacopilot}; TRIANGLE and FLASH study triage~\cite{triangle2025,flair2025,zhu2025decolifecyclemanagemententerprisegrade}. Cloud-OpsBench~\cite{wang2026cloudopsbench} proposes a reproducible evaluation framework for AIOps agents that accounts for analysis trajectory quality, while Trace2Skill~\cite{ni2026trace2skill} automates skill creation and update from agent trajectories using multi-agent distillation. Neuro-symbolic approaches use structured graphs to constrain LLM outputs~\cite{delong2024neurosymbolic}. Tool-augmented agents and agentic RAG support iterative reasoning~\cite{yao2023react,schick2023toolformer,singh2025agenticrag,asai2024selfrag}. However, these systems either rely on manually curated knowledge artifacts, learn flat skill libraries without causal structure, or treat the knowledge base as static after construction. \sysname bridges these gaps by grounding agentic traversal in explicit problem--action relationships within a self-evolving workflow graph, which can in turn serve as the knowledge backbone for the aforementioned systems.

%% file: 3-offline_construction.tex
\section{Pillar 1: Graph Construction (Warm Start)}
\label{sec:offline}

\begin{figure}[t]
    \centering
    \includegraphics[width=\columnwidth]{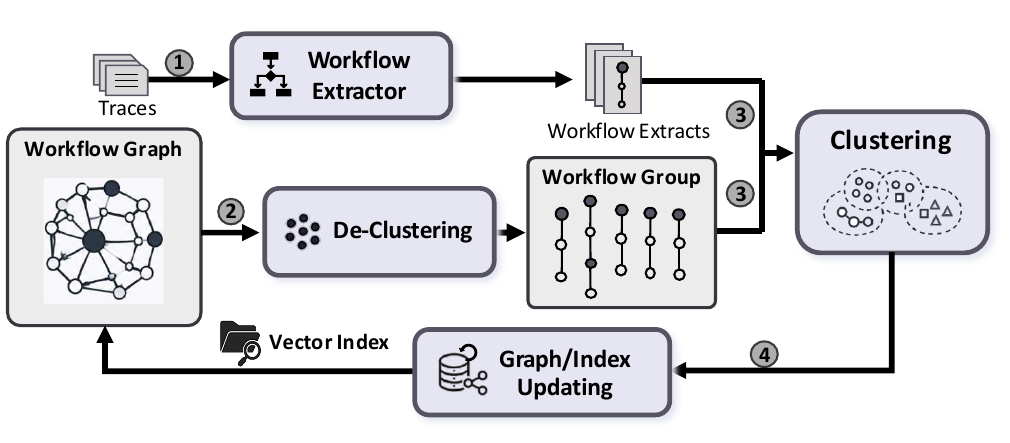}
    \vspace{-0.6cm}
    \caption{Incremental graph construction: new traces are extracted, merged with the de-clustered existing graph, and re-clustered to produce a deduplicated update.}
    \label{fig:offline_e2e}
    \vspace{-0.4cm}
\end{figure}

The offline pipeline transforms raw operational traces into a structured workflow graph. 
The pipeline supports \emph{incremental construction}, integrating new traces without rebuilding the graph.
As illustrated in Figure~\ref{fig:offline_e2e}, construction proceeds in four steps:
\emph{(1)~Workflow Extraction}: an LLM-based extractor produces one workflow extract per trace.
\emph{(2)~De-clustering and Merging}: clustered nodes are reverted to raw extracts and merged with newly extracted ones.
\emph{(3)~Clustering}: semantically similar nodes are identified and merged, deduplicating with edges rewired.
\emph{(4)~Graph and Index Update}: the graph is synchronized to the cloud database and vector index.

\subsection{Workflow Extractor}
\label{sec:extraction}

The graph uses three entity types: \textbf{Domain ($\mathcal{D}$)} (bounded operational area), \textbf{Problem ($\mathcal{P}$)} (issue requiring resolution), and \textbf{Action ($\mathcal{A}$)} (executable step), connected by four edge types: \texttt{CAUSES} ($p_i \!\to\! p_j$), \texttt{RESOLVES} ($a \!\to\! p$), \texttt{LEADS\_TO} ($a_i \!\to\! a_j$, sequential dependency), and \texttt{BELONGS\_TO} (domain membership). This yields $\mathcal{G} = (\mathcal{V}, \mathcal{E})$ where $\mathcal{V} = \mathcal{D} \cup \mathcal{P} \cup \mathcal{A}$, supporting structured traversal from root causes to resolutions.


A \textbf{multi-agent extraction pipeline} ingests heterogeneous sources (issue-tracking databases, historical resolution traces) and proceeds in three stages: \emph{Stage~1 (Preprocessing)} normalizes formats, converts images/tables to text via vision-language models, and segments documents; \emph{Stage~2 (Construction)} extracts problem statements, identifies discrete action steps with preconditions, and assembles reasoning chains conforming to the graph schema; \emph{Stage~3 (Pruning)} removes redundant nodes and verifies type consistency. All stages are LLM-driven using the graph schema as part of the system prompt.


\subsection{Clustering}
\label{sec:clustering}
As workflows accumulate, the graph acquires semantically similar nodes. Instead of using semantically rich LLM-based deduplication, we apply efficient embedding-based agglomerative clustering~\cite{johnson1967hierarchical} separately for each node type: (1)~node text is canonicalized via regex masking to remove trace-specific tokens (GUIDs, timestamps, IPs); (2)~canonicalized descriptions are embedded (e.g., \texttt{text-embedding-3-large}); (3)~agglomerative clustering with cosine distance threshold $\tau$ groups similar nodes; (4)~for each cluster, a representative node $n_C^*$ is generated from three members (longest, shortest, median), inheriting the \textit{union} of all member edges; (5)~original members are flagged as disabled and excluded from retrieval.

\subsection{De-Clustering and Maintenance}
\label{sec:update}
When new traces arrive, existing cluster representatives are removed and all original members re-enabled (de-clustering). Raw workflows are merged with newly extracted ones, and clustering is re-applied to produce the updated graph, maintaining deduplication while supporting incremental updates without full reconstruction.

%% file: 4-online.tex
\section{Pillar 2: Multi-Agent Graph Traversal}
\label{sec:online}

Each online traversal corresponds to a single autonomous execution: a team of agents iteratively navigates the workflow graph, selecting edges and nodes based on a combination of reinforcement weights (encoding historical reward signals) and LLM-driven reasoning, executing the chosen actions, and producing a trajectory that serves as input to the ATR layer (Section~\ref{sec:aco}).

\subsection{Orchestration}
We leverage an existings orchestration framework~\cite{zhu2025decolifecyclemanagemententerprisegrade} as the online agentic execution environment; the orchestration layer is modular and replaceable by any Model Context Protocol (MCP)-compatible client~\cite{mcp} (e.g., GitHub Copilot CLI~\cite{githubcopilotcli}, Claude Code~\cite{claudecode}). The framework employs a nested control flow: an \emph{outer loop} over \textbf{agents} and an \emph{inner loop} over \textbf{tools}. Each outer iteration maintains a \textbf{global context} $C_t$ (long-term memory); each tool writes to a \textbf{temporary context} $c_{t,j}$ (short-term memory). The \texttt{Coordination Agent} drives reasoning via \texttt{Graph Loader} and \texttt{Action Planner} tools, while \texttt{Execution Agents} (e.g., \texttt{Kusto Agent}) use tools to execute actions and return results for iterative refinement.

\subsection{Agentic Graph-Traversal Algorithm}

The learned workflow graph can contain tens of thousands of nodes and edges.
Given this scale, the online execution engine requires a traversal algorithm that is \textbf{efficient, robust, and accurate}, so it can identify the most relevant next actions based on principled reasoning, reinforcement-weighted edge priorities, and on the outcomes of previously executed actions. Given workflow graph $\mathcal{G} = (\mathcal{V}, \mathcal{E})$ that consists of typed nodes (domains, problems, actions) connected by causal and procedural edges, we organize the traversal as a nested control flow (Algorithm~\ref{alg:e2e-agentic-graph-rag}). The outer loop (index $t$) corresponds to major investigation iterations with context $C_t$; the inner loop (index $j$) performs iterative graph exploration with the following tools exposed to the agent:

\smallskip\noindent\textbf{Graph Loader.}
At each inner-loop step $(t,j)$ with context $c_{t,j}$, the \texttt{Tool Selector} dynamically determines a graph query text $s_{t,j}$ and a target node type $\tau_{t,j}$ as part of the tool invocation argument. This step requires strong LLM reasoning: much like an MCP client~\cite{mcp} or function-calling agent~\cite{openai2023functioncalling} that must reason about tool semantics to determine the correct arguments, $f_{\text{LLM}}$ must interpret the current investigation context, decide which type of graph node is most relevant, and formulate a precise query to retrieve it:
\begin{align}
(s_{t,j}, \tau_{t,j}) \gets f_{\text{LLM}}(C_t, c_{t,j}).
\end{align}
The \texttt{Graph Loader} then retrieves a set of top-$k_p$ root nodes $\mathcal{R}_{t,j}$ of type $\tau_{t,j}$ using vector similarity search with low latency given the pre-built search index:
\begin{align}
q_{t,j} &\leftarrow \operatorname{Embed}(s_{t,j}), \\
\mathcal{R}_{t,j} &\leftarrow \operatorname{Top}_{k_p}\big(
    \mathcal{V}_{\tau_{t,j}} \setminus \mathcal{R}_{t,<j},\; q_{t,j}
\big),
\end{align}
where $\mathcal{V}_{\tau_{t,j}} \subseteq \mathcal{V}$ is the set of nodes of type $\tau_{t,j}$, and $\mathcal{R}_{t,<j} = \bigcup_{j' < j} \mathcal{R}_{t,j'}$ denotes root nodes previously retrieved within the same outer-loop iteration $t$. This exclusion encourages exploration diversity by directing each re-invocation toward a different region of the graph.
The loader then expands from $\mathcal{R}_{t,j}$ to extract an $m$-hop neighborhood subgraph via \emph{probabilistic multi-path expansion}: at each hop, outgoing edges are sampled without replacement with probability proportional to their log-compressed reinforcement weights $\log(1+\phi(e))$ (detailed in Section~\ref{sec:pheromone_selection}). This yields the expanded subgraph:
\begin{align}
\mathcal{V}_{t,j} &\leftarrow \{ v \in \mathcal{V} \mid \exists r \in \mathcal{R}_{t,j},\ v \text{ reached within } m \text{ hops} \}, \\
\mathcal{E}_{t,j} &\leftarrow \{ (u,v) \in \mathcal{E} \mid u \in \mathcal{V}_{t,j},\ v \in \mathcal{V}_{t,j} \}, \notag \\
\mathcal{G}_{t,j} &\leftarrow (\mathcal{V}_{t,j},\mathcal{E}_{t,j}). \notag
\end{align}
This probabilistic expansion biases the subgraph toward historically successful neighborhoods while still allowing less-traversed paths to be discovered.

\smallskip\noindent\textbf{Action Planner.}
The \texttt{Action Planner} picks actions by conditioning an LLM on the investigation context, the extracted subgraph (with node reinforcement weights $\phi(v)$ annotated), and system-level instructions:
\begin{align}
(A_{t,j}, \Delta c_{t,j}) &\gets g_{\text{LLM}}(C_t, c_{t,j}, \mathcal{G}_{t,j}, \{\phi(v)\}_{v \in \mathcal{V}_{t,j}}),
\end{align}
where $A_{t,j}$ contains up to $k_a$ selected actions and the LLM output $\Delta c_{t,j}$ captures auxiliary reasoning such as new hypotheses, observations, or suggested exploration directions. If $A_{t,j} = \varnothing$, the planner has not found actionable steps in the current graph neighborhood; in this case, $\Delta c_{t,j}$ serves as a pivoting signal containing hints about alternative hypotheses or unexplored problem areas. This context update is merged into the temporary context: $c_{t,j+1} \leftarrow c_{t,j} \cup \Delta c_{t,j}$, so that in the next inner-loop iteration the enriched $c_{t,j+1}$ guides the \texttt{Graph Loader} to search a different region of the graph, effectively steering exploration toward more promising neighborhoods. 

\smallskip\noindent\textbf{Action Execution.}
Once the \texttt{Action Planner} produces a non-empty set $A_{t,j}$, the corresponding \texttt{Execution Agents} take over. For each selected action $a \in A_{t,j}$, a specialized agent (e.g., \texttt{Kusto Agent} for diagnostic queries) executes the action and returns an observation $o_t^{(a)} \gets \operatorname{Exec}(a)$. The observation is then incorporated into the global investigation context: $C_t \gets C_t \cup \{(a, o_t^{(a)})\}$. After all actions in $A_{t,j}$ have been executed, the updated context $C_{t+1} \gets C_t$ is carried forward to the next outer-loop iteration (Algorithm~\ref{alg:e2e-agentic-graph-rag}, lines 14), enabling subsequent iterations to reason over accumulated evidence from all prior actions. When all iterations complete and no further actions are recommended, the system reaches a termination state; an LLM then synthesizes the full accumulated context $C_t$ into a final workflow execution report (Algorithm~\ref{alg:e2e-agentic-graph-rag}, line 16).

\begin{algorithm}[t]
\caption{Agentic Graph-Traversal with Nested Loops}
\label{alg:e2e-agentic-graph-rag}
\small
\begin{algorithmic}[1]
\Require Graph $\mathcal{G}=(\mathcal{V},\mathcal{E})$, context $C_0$, max inner iterations $J$
\State $t \gets 0$
\While{not terminated}
    \State $j \gets 0;\; c_{t,0} \gets \emptyset;\; A_{t,j} \gets \varnothing$
    \While{$A_{t,j} = \varnothing$ \textbf{and} $j < J$}
        \State $\mathcal{G}_{t,j} \gets \textsc{GraphLoader}(C_t, c_{t,j}, \mathcal{G})$
        \State $(A_{t,j}, \Delta c_{t,j}) \gets \textsc{ActionPlanner}(C_t, c_{t,j}, \mathcal{G}_{t,j})$
        \State $c_{t,j+1} \gets c_{t,j} \cup \Delta c_{t,j}$;\; $j \gets j+1$
    \EndWhile
    \If{$A_{t,j-1}=\varnothing$} \textbf{terminate} \EndIf
    \ForAll{$a \in A_{t,j-1}$}
        \State $o_t^{(a)} \gets \operatorname{Exec}(a)$;\; $C_t \gets C_t \cup \{(a, o_t^{(a)})\}$
    \EndFor
    \State $C_{t+1} \gets C_t$;\; $t \gets t+1$
\EndWhile
\State \Return Final workflow execution report
\end{algorithmic}
\end{algorithm}

%


%% file: 4a-aco_evolution.tex
\section{Pillar 3: Adaptive Traversal Reinforcement}
\label{sec:aco}

The ATR layer closes the loop between offline construction and online execution by treating each traversal as a reinforcement signal.

\subsection{Reinforcement Model}
\label{sec:pheromone_model}

We augment $\mathcal{G} = (\mathcal{V}, \mathcal{E})$ with a reinforcement function $\phi: \mathcal{E} \cup \mathcal{V} \to \mathbb{R}^+$, initialized uniformly at $\phi_0$.

\subsection{Reinforcement Deposition}
\label{sec:deposition}


Every online traversal produces a \emph{trajectory} 
$\mathcal{T}=\langle (v_i,e_i,o_i)\rangle_{i=1}^{n}$, recording the visited nodes, traversed edges, and execution observations. The system then assigns a \emph{solution quality score} $Q(\mathcal{T})\in[0,1]$:
\begin{align}
\resizebox{.91\columnwidth}{!}{$
Q(\mathcal{T}) =
w_{\text{useful}}\cdot \mathrm{usefulness}(\mathcal{T})
+ w_{\text{gnd}}\cdot \mathrm{groundedness}(\mathcal{T})
+ w_{\text{usr}}\cdot \mathrm{user\_score}(\mathcal{T})
$},
\end{align}
where usefulness is LLM-judged, groundedness measures evidence support, and user feedback, when available, takes precedence. Other scoring functions can be used depending on the scenario.

Reinforcement is deposited only when quality reaches threshold $\delta_q$ (set to $0.8$), normalized by trajectory length to bias toward efficient sequences:
\begin{align}
    \Delta\phi(x) = \begin{cases} \frac{Q(\mathcal{T})}{|\mathcal{T}|}, & \text{if } Q(\mathcal{T}) \geq \delta_q \\ 0, & \text{otherwise} \end{cases}, \quad \forall x \in \{e_i, v_i\} \subseteq \mathcal{T}
    \label{eq:deposition}
\end{align}

\subsection{Temporal Decay}
\label{sec:evaporation}

All weights are decayed at regular intervals: $\phi(x) \gets (1 - \rho) \cdot \phi(x)$, $\forall x \in \mathcal{E} \cup \mathcal{V}$, where $\rho \in (0, 1)$ controls the memory horizon. Decay removes stale reinforcement, prevents premature convergence, and enables adaptation to infrastructure changes.

\subsection{Reinforcement-Weighted Subgraph Retrieval}
\label{sec:pheromone_selection}

During online traversal, the \texttt{Graph Loader} uses reinforcement weights to probabilistically expand the subgraph. At node $u$, outgoing edges are sampled with the following probability:
\begin{align}
    p(e) = \frac{\bigl[\log(1 + \phi(e))\bigr]^\alpha}{\sum_{e' \in \mathcal{N}(u)} \bigl[\log(1 + \phi(e'))\bigr]^\alpha},
    \label{eq:edge_select}
\end{align}
where $\alpha$ interpolates between exploitation ($\alpha \!\to\! \infty$) and exploration ($\alpha \!\to\! 0$). Node weights $\phi(v)$ serve as soft priors in the \texttt{Action Planner}'s context. Additionally, when a successful traversal consecutively visits unlinked nodes, a new \texttt{LEADS\_TO} edge is synthesized, capturing cross-incident sequences discovered online.

%% file: 5-evaluation.tex
\section{Evaluation}
We evaluate \sysname in three settings: (i)~offline extraction quality (Section~\ref{sec:offline_eval}); (ii)~controlled online evaluation on held-out incidents (Section~\ref{sec:online_eval}); and (iii)~ATR impact on graph evolution and traversal (Section~\ref{sec:aco_eval}). 

%% file: 5a-offline_eval.tex
\subsection{Offline Graph Construction Evaluation}
\label{sec:offline_eval}

To evaluate the graph construction accuracy, we invited domain experts to annotate 30 incident tickets (71 problem nodes, 191 action nodes, 262 edges) as ground truth.

\paragraph{Extraction Quality.}
Table~\ref{tab:node_extraction} summarizes results. The pipeline achieves node F1=0.93 (96\% precision, 90\% recall) and edge F1=0.92. Among edge types, \texttt{CAUSES} achieves F1=0.98, \texttt{LEADS\_TO} F1=0.91 (implicit ordering is challenging), and \texttt{RESOLVES} F1=0.86 (linking actions to outcomes is hardest).

\begin{table}[t]
    \centering
    \caption{Node (left) and edge (right) extraction accuracy.}
        \vspace{-0.2cm}
    \label{tab:node_extraction}
    \label{tab:edge_extraction}
    \resizebox{0.9\columnwidth}{!}{%
    \begin{tabular}{lccc}
        \toprule
        \textbf{Node Type} & \textbf{Prec.} & \textbf{Rec.} & \textbf{F1} \\
        \midrule
        Problem Nodes & 0.98 & 0.92 & 0.95 \\
        Action Nodes & 0.96 & 0.90 & 0.92 \\
        \midrule
        \textbf{Overall} & \textbf{0.96} & \textbf{0.90} & \textbf{0.93} \\
        \bottomrule
    \end{tabular}%
    \quad
    \begin{tabular}{lccc}
        \toprule
        \textbf{Edge Type} & \textbf{Prec.} & \textbf{Rec.} & \textbf{F1} \\
        \midrule
        \texttt{CAUSES} & 1.00 & 0.95 & 0.98 \\
        \texttt{RESOLVES} & 0.86 & 0.86 & 0.86 \\
        \texttt{LEADS\_TO} & 0.90 & 0.93 & 0.91 \\
        \midrule
        \textbf{Overall} & \textbf{0.91} & \textbf{0.92} & \textbf{0.92} \\
        \bottomrule
    \end{tabular}%
    }
    \vspace{-0.5cm}
\end{table}



\paragraph{Error Analysis.}
Manual inspection of extraction errors reveals four categories: implicit actions not surfaced as explicit nodes (38\%), domain jargon causing misclassification (26\%), unresolved cross-references to external documents or prior incidents (21\%), and incomplete multimodal fusion across text and images (15\%).

\begin{figure}[t]
    \centering
    \begin{subfigure}[b]{0.48\columnwidth}
        \centering
        \includegraphics[width=\textwidth]{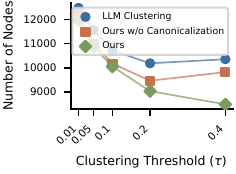}
        \caption{Node count vs.\ threshold.}
        \label{fig:graph_size_nodes}
    \end{subfigure}
    \hfill
    \begin{subfigure}[b]{0.48\columnwidth}
        \centering
        \includegraphics[width=\textwidth]{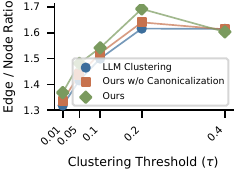}
        \caption{Edge/node ratio vs.\ threshold.}
        \label{fig:graph_size_ratio}
    \end{subfigure}
    \vspace{-0.2cm}
    \caption{Graph size under varying clustering thresholds.}
    \label{fig:graph_size_comparison}
    \vspace{-0.4cm}
\end{figure}

\paragraph{Clustering Analysis.}
We evaluate clustering quality across thresholds 0.01--0.4 with three variants: \emph{Ours} (canonicalization + embedding clustering), \emph{w/o canonicalization} (embedding only), and \emph{LLM Clustering} (for each embedding-based cluster, an LLM produces the final cluster assignments to capture richer semantic duplication). Higher thresholds reduce graph size (from ${\sim}$12K to ${\sim}$8.5--10K nodes) while increasing edge-to-node ratio from 1.3 to 1.6, yielding denser graphs (Figure~\ref{fig:graph_size_comparison}). Our canonicalization variant achieves comparable reduction to LLM Clustering at significantly lower cost.

\paragraph{Cost and Latency.}
Average cost is \$0.22/incident (15.7 LLM calls, ${\sim}$54K tokens); median latency 1.3 min (mean 2.9 min).

%% file: 5b-online_eval.tex
\subsection{Online Traversal Evaluation}
\label{sec:online_eval}

We evaluate \sysname's agentic traversal using the graph constructed from one year of human resolution traces spanning 1{,}330 production incidents resolved before February 2026, with 13{,}344 active nodes at clustering threshold $\tau{=}0.01$. Test incidents are drawn exclusively from tickets filed \emph{after} this date, ensuring a \textbf{strict temporal split} with no information leakage. All LLM components use GPT-5.2.
Each test incident is investigated by \sysname autonomously without human intervention: given only the incident description, the system iteratively traverses the workflow graph, executes diagnostic actions (e.g., KQL telemetry queries), and attempts to reach a resolution.
We report four aggregate metrics: (1)~\textit{Mitigation Reach}, the fraction of incidents where the system completes its investigation with a conclusive output rather than halting prematurely; (2)~\textit{Hallucination Rate}, the fraction of conclusions not grounded in executed diagnostic evidence; (3)~\textit{Usefulness}, an LLM-judged 1--5 score comparing recommended actions against the human resolution log; and (4)~\textit{KQL Successes}, the number of KQL queries returning non-empty results per incident, because incorrectly targeted queries typically fail or return empty results.



\subsubsection*{\sysname vs.\ Agentic Summary-RAG}\label{sec:rag_comparison}

We compare \textbf{\sysname} against an \textbf{Agentic Summary-RAG} baseline that retrieves LLM-condensed summaries of past incident traces via similarity search and feeds them to an LLM agent orchestrator that decides when and how to perform retrieval iteratively, following the agentic RAG paradigm~\cite{singh2025agenticrag}. Both systems use the same agentic orchestration loop and benefit from LLM-based preprocessing; the key difference is knowledge representation: \textbf{Agentic Summary-RAG} retrieves per-incident summaries as flat documents, whereas \textbf{\sysname} retrieves from a deduplicated workflow graph with explicit causal and procedural edges.

As a qualitative validation, two domain experts conducted blind reviews of 19 matched incidents (the maximum feasible due to expert-availability constraints). \textbf{\sysname} scored 4.95/5 (94.7\% perfect) vs.\ 3.68 for \textbf{Agentic Summary-RAG} (26.3\% perfect). While the sample is small, the effect size is large (Cohen's $d{>}2$) and consistent: experts noted that \textbf{Agentic Summary-RAG} frequently produced incorrect query predicates and incomplete investigations.

\begin{table}[t]
    \centering
    \caption{Comparison of \textbf{\sysname} against \textbf{Agentic Summary-RAG}. Values are mean$\,\pm\,$95\% CI over 4 runs (2 per method). Best in \textbf{bold}.}
    \label{tab:rag_comparison}
        \vspace{-0.1cm}
    \resizebox{\columnwidth}{!}{%
    \begin{tabular}{lcccc}
        \toprule
        \textbf{Method} & \textbf{Mit.\ Reach}$\uparrow$ & \textbf{Halluc.\ Rate}$\downarrow$ & \textbf{Useful.}$\uparrow$ & \textbf{KQL Succ.}$\uparrow$ \\
        \midrule
        \textbf{Agentic Summary-RAG}& $90.9 \pm 6.1$\% & $22.7 \pm 8.9$\%          & $3.32 \pm 0.20$          & $2.5 \pm 0.4$           \\
        \textbf{\sysname} ($\tau{=}0.01$)    & $\mathbf{98.5} \pm 2.1$\% & $\mathbf{3.8} \pm 3.3$\% & $\mathbf{3.49} \pm 0.12$ & $\mathbf{10.0} \pm 1.4$ \\
        \bottomrule
    \end{tabular}%
    }
\end{table}

\begin{figure}[t]
    \centering
    \begin{subfigure}[b]{0.48\columnwidth}
        \centering
        \includegraphics[width=\textwidth]{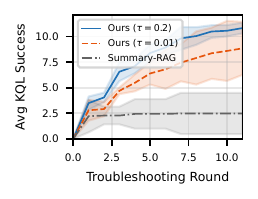}
        \vspace{-0.6cm}
        \caption{Cumul.\ KQL success.}
        \label{fig:kql_graph_vs_rag}
    \end{subfigure}
    \hfill
    \begin{subfigure}[b]{0.48\columnwidth}
        \centering
        \includegraphics[width=\textwidth]{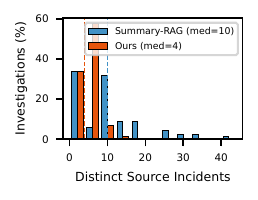}
        \vspace{-0.6cm}
        \caption{Source incident dist.}
        \label{fig:rag_ticket_hist}
    \end{subfigure}
    \caption{(a)~Cumulative KQL success per troubleshooting round. (b)~Source incidents referenced per investigation.}
    \label{fig:rag_comparison}
    \vspace{-0.3cm}
\end{figure}

As a complement to the human review, table~\ref{tab:rag_comparison} reports automated metrics on 93 held-out incidents. \textbf{\sysname} outperforms on mitigation reach (98.5\% vs.\ 90.9\%), hallucination rate (3.8\% vs.\ 22.7\%), and usefulness (3.49 vs.\ 3.32), with 4$\times$ higher KQL throughput (10.0 vs.\ 2.5; Figure~\ref{fig:kql_graph_vs_rag}).
Figure~\ref{fig:rag_ticket_hist} shows \textbf{Agentic Summary-RAG} references a median of 10 source incidents vs.\ only 4 for \textbf{\sysname}, yet achieves lower quality. This gap reflects a fundamental difference in context efficiency: \textbf{Agentic Summary-RAG}'s first retrieval injects a mean of ${\sim}$32K tokens of per-incident summaries into the agent context, whereas \textbf{\sysname}'s first graph load provides a mean of ${\sim}$3.9K tokens, an \textbf{8$\times$ reduction} in retrieval context. 

\smallskip
\subsubsection*{Why Not Raw-Trace Retrieval?}\label{sec:raw_rag}
We additionally compare three formats on the knowledge-compression spectrum: \textbf{Trace-RAG}, LLM-condensed summaries (\textbf{Agentic Summary-RAG}), and \textbf{\sysname}.
Raw incident logs average ${\sim}$6.5K tokens (P95: ${\sim}$27K; max: ${\sim}$47K). With \textbf{Agentic Summary-RAG} retrieving a median of 10 source incidents (19\% retrieving more than 10), \textbf{Trace-RAG} would consume ${\sim}$65K tokens on average (51\% of a 128K context window), with worst cases exceeding 315K tokens. Even when raw logs fit within the context window, they contain substantial noise (automated enrichment messages, status updates, formatting artifacts) that degrades signal-to-noise ratio and increases hallucination. This motivates both \textbf{Agentic Summary-RAG}'s LLM-condensed summaries and \textbf{\sysname}'s graph-based deduplication as necessary compression, with \textbf{\sysname} achieving further savings through cross-incident node connections.

\subsubsection*{Orchestration Comparison: Native vs.\ GHC}
\label{sec:copilot_comparison}

We compare \textbf{\sysname}'s \textbf{native} orchestrator (Section~\ref{sec:online}) against \textbf{GHC} (GitHub Copilot~\cite{githubcopilotcli}), a general-purpose agentic framework that accesses the same graph and tools via Model Context Protocol (MCP)~\cite{mcp}. Note that GHC is not configured for iterative graph traversal; this comparison therefore measures how well a general-purpose orchestrator utilizes \sysname's graph, rather than a controlled baseline comparison.
\textbf{GHC} achieves 42.0\% hallucination rate vs.\ the \textbf{native} orchestrator's 3.8\%, and only 2.4 KQL queries per incident vs.\ 10.0. GHC adopts a plan-once-then-execute strategy rather than iterative explore--exploit--replan cycles, yielding shallower investigations despite spending its budget on reasoning (8.1 steps/ticket) and tool discovery (2.6 tool lookups). This gap highlights the importance of structured orchestration for effective graph utilization.


\subsubsection*{Sensitivity Analysis: Retrieval Parameters}
\begin{figure}[t]
    \centering
    \includegraphics[width=\columnwidth]{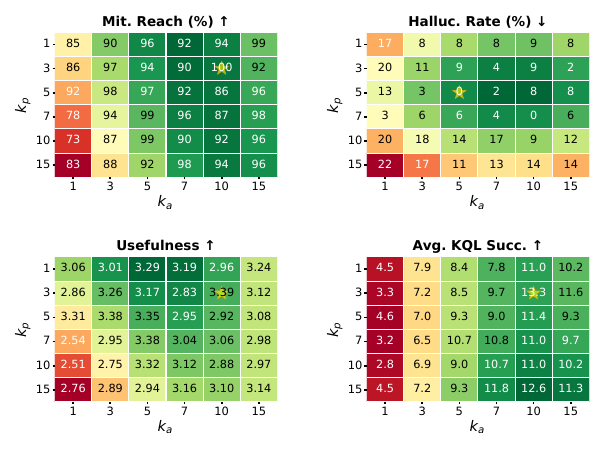}
    \vspace{-0.9cm}
    \caption{Impact of $(k_p, k_a)$. Best cell marked with $\ast$.}
    \label{fig:heatmap_kp_ka}
        \vspace{-0.5cm}
\end{figure}

We study the impact of $k_p$, the number of root nodes, and $k_a$, the number of action nodes selected per round, on online performance across a $6{\times}6$ grid with $k_p,k_a \in \{1,3,5,7,10,15\}$ (3 runs each; Figure~\ref{fig:heatmap_kp_ka}).
Increasing $k_a$ improves KQL throughput, but $k_a{=}1$ is significantly worse across all metrics, confirming that action diversity is critical.
The $k_p$ parameter shows a concave pattern: moderate values ($k_p{\in}\{3,5\}$) yield the best mitigation reach and lowest hallucination rate, while extremes underperform. The $(k_p{=}3, k_a{=}7)$ region achieves the best balance.

\subsubsection*{Cross-Domain Generalizability}
To evaluate whether \textbf{\sysname} generalizes beyond the incident investigation scenario, we applied the same architecture to a completely different domain: SpreadsheetBench~\cite{ma2024spreadsheetbench}, a spreadsheet formula benchmark (124 held-out samples, 6 runs). \textbf{\sysname} achieved $19.1{\pm}1.9$\% Pass@1 compared to $17.2{\pm}1.5$\% for \textbf{Trace-RAG}.

%% file: 5c-aco_eval_v2.tex
\subsection{Adaptive Traversal Reinforcement Evaluation}
\label{sec:aco_eval}

We evaluate ATR along two dimensions: graph evolution behavior and impact on online traversal.

\subsubsection*{Graph Evolution}
\label{sec:graph_evolution}

\begin{figure}[t]
    \centering
    \begin{subfigure}[b]{0.48\columnwidth}
        \centering
        \includegraphics[width=\textwidth]{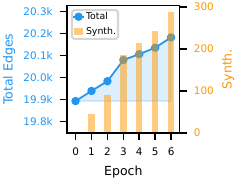}
        \vspace{-0.5cm}
        \caption{Cumul.\ edge synthesis.}
        \label{fig:edge_synthesis}
    \end{subfigure}
    \hfill
    \begin{subfigure}[b]{0.48\columnwidth}
        \centering
        \includegraphics[width=\columnwidth]{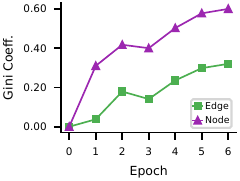}
        \vspace{-0.5cm}
        \caption{Gini coefficient.}
        \label{fig:gini_convergence}
    \end{subfigure}
    \vspace{-0.2cm}
    \caption{Reinforcement evolution over six epochs.}
    \label{fig:pheromone_convergence}
\end{figure}

We evolve the workflow graph over six reinforcement epochs (15 traversal trajectories each). We use the following configuration: decay $\rho{=}0$ (retaining all reinforcement) and concentration $\alpha{=}0$ (uniform edge selection), isolating the effect of reinforcement deposition and edge synthesis\footnote{At the current graph scale, the full $m$-hop subgraph retrieved from each root node still fits within the 1M-token context window. Reinforcement-based trimming would further improve token efficiency.}. The graph begins with 13{,}344 nodes and 19{,}893 edges. After six epochs, 289 new edges are synthesized as traversals discover cross-incident action sequences, growing the total to 20{,}182. Figure~\ref{fig:pheromone_convergence}(a) shows cumulative edge synthesis growth; Figure~\ref{fig:pheromone_convergence}(b) tracks the Gini coefficient~\cite{gini1921measurement} of reinforcement weights: node Gini rises to 0.60 and edge Gini to 0.32, indicating concentration on frequently successful paths while maintaining a long tail of alternatives.

\begin{figure}[t]
    \centering
    \includegraphics[width=\columnwidth]{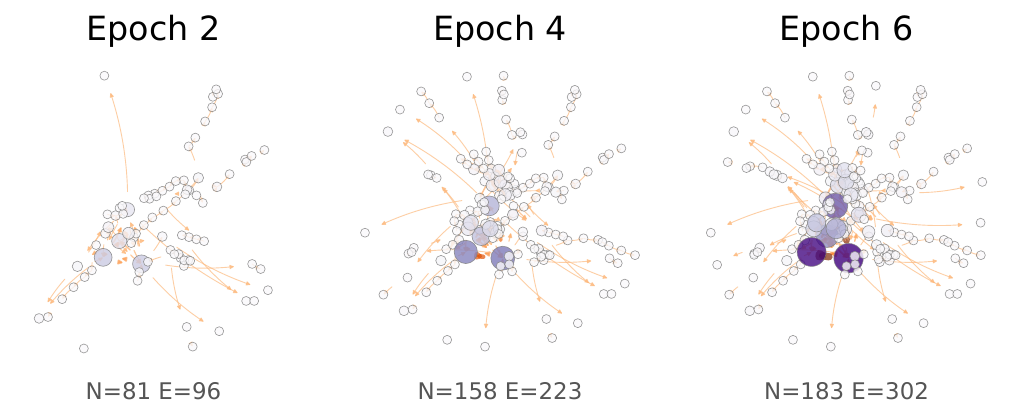}
    \vspace{-0.6cm}
    \caption{Reinforced subgraph evolution at epochs 2, 4, and 6. Node size and edge thickness are proportional to reinforcement weight. ``Highway'' clusters emerge as traversal experience accumulates.}
    \label{fig:graph_evolution}
\end{figure}

Figure~\ref{fig:graph_evolution} visualizes the reinforced subgraph at epochs 2, 4, and 6. Distinct ``highway'' clusters emerge for common failure categories (e.g., capacity, connectivity, authentication), connected by synthesized edges that bridge incident types and enable cross-domain diagnostic transfer.

\subsubsection*{Impact on Online Traversal}
\label{sec:aco_efficiency}

We compare \emph{With Reinforcement} (learned weights $\phi$ bias edge selection ) against \emph{Without Reinforcement} (uniform weights) on 163 recent incidents across 8 runs (4 per condition)\footnote{The 98.5\% mitigation reach in Section~\ref{sec:rag_comparison} is from a 93-incident evaluation. The 92--93\% here reflects a different set of 163 incidents: the ATR ablation was conducted at a later time, and the available incidents differ because each evaluation counts back from its run date over the telemetry data retention window.}. Notably, evaluation incidents may involve different problem categories than those encountered during deposition, testing whether reinforcement generalizes beyond the deposited paths.

\begin{table}[t]
    \centering
    \caption{Impact of ATR on online performance over 8 runs (4 per condition). Best in \textbf{bold}.}
    \label{tab:aco_ablation}
    \resizebox{\columnwidth}{!}{%
    \begin{tabular}{lcccc}
        \toprule
        \textbf{Condition} & \textbf{Mit.\ Reach}$\uparrow$ & \textbf{Halluc.\ Rate}$\downarrow$ & \textbf{Useful.}$\uparrow$ & \textbf{KQL Succ.}$\uparrow$ \\
        \midrule
        \textbf{W/o Reinf.} & $92.0 \pm 2.1$\% & $10.5 \pm 2.3$\% & $3.331 \pm 0.060$ & $15.1 \pm 1.1$ \\
        \textbf{W/ Reinf.}  & $\mathbf{93.3} \pm 1.9$\% & $\mathbf{7.8} \pm 2.1$\% & $\mathbf{3.366} \pm 0.061$ & $\mathbf{17.2} \pm 1.2$ \\
        \bottomrule
    \end{tabular}%
    }
\end{table}

Table~\ref{tab:aco_ablation} shows that reinforcement reduces hallucination rate by 26\% ($7.8\%$ vs.\ $10.5\%$) and improves mitigation reach slightly ($93.3\%$ vs.\ $92.0\%$), while KQL throughput increases from 15.1 to 17.2. Although confidence intervals overlap for some individual metrics, improvements are \emph{directionally consistent} across all four metrics and all four independent runs per condition, indicating a robust signal rather than statistical noise. However, \textit{this accuracy gain is not free}: reinforced traversal explores more paths and executes more actions, resulting in approximately 8\% higher end-to-end latency on average and on average 1.5 more rounds of investigation.

%% file: 6-deployment.tex
\section{Deployment and Usage Study}
\label{sec:deployment}

We report a field study of \sysname's production deployment across four cloud database services at \company, covering 194 real-world investigation conversations over a 12-week period using two orchestrators: the native orchestrator (143 conversations) and GHC (51 conversations).
All conversations were initiated by on-call engineers during live incident triage.
Of the 143 native orchestrator conversations, 62 were independently evaluated by domain experts on a 1--5 usefulness scale.

\subsubsection*{Adoption and Engagement.}
\begin{figure}[t]
    \centering
    \begin{subfigure}[b]{0.48\columnwidth}
        \centering
        \includegraphics[width=\textwidth]{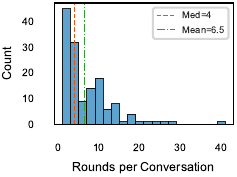}
        \vspace{-0.5cm}
        \caption{Conversation depth.}
        \label{fig:deploy_depth}
    \end{subfigure}
    \hfill
    \begin{subfigure}[b]{0.48\columnwidth}
        \centering
        \includegraphics[width=\textwidth]{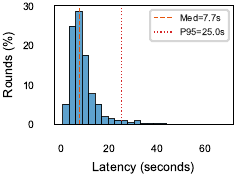}
        \vspace{-0.5cm}
        \caption{Per-round latency.}
        \label{fig:deploy_latency}
    \end{subfigure}
    \vspace{-0.2cm}
    \caption{(a)~Conversation depth distribution (median 4 rounds). (b)~Per-round latency (median 7.7\,s).}
    \label{fig:deploy_depth_latency}
    \vspace{-0.3cm}
\end{figure}

Usage grew substantially over the study period, with 95 of the 143 native orchestrator conversations (66\%) occurring during the final four weeks. Figure~\ref{fig:deploy_depth_latency} summarizes all 143 conversations. The median conversation spans 4 rounds (mean~6.5), with a right-skewed tail reaching up to 40 rounds for complex incidents.
Median per-round latency is 7.7\,s (P95: 25.0\,s), remaining within interactive bounds despite multi-step graph retrieval, LLM planning, KQL execution, and result synthesis.

\subsubsection*{Diagnostic Effectiveness and Outcomes.}
The system generated 1{,}299 KQL queries across 143 conversations, of which 1{,}077 (83\%) returned data (median 6 per conversation).
Among the 62 scored conversations, manual outcome classification shows that 26\% of conversations ($n{=}16$) reached a definitive root cause, 71\% ($n{=}44$) produced useful diagnostic evidence advancing the investigation, and only 3\% ($n{=}2$) yielded no actionable output.
The mean usefulness score is 3.10/5, with 79\% rated $\geq$3 and 34\% rated 4--5.

\smallskip\noindent\textbf{Failure Analysis.}
We analyzed the 21\% of conversations scoring below 3 and the 3\% ($n{=}2$) with no actionable output. Three dominant failure modes emerge: (1)~\emph{coverage gaps} (45\% of failures): incidents involving recently deployed services or rare failure modes not yet represented in the graph; (2)~\emph{telemetry expiration} (35\%): diagnostic queries returning empty results because retention windows had elapsed by investigation time; and (3)~\emph{cross-team dependencies} (20\%): incidents requiring access to restricted systems or data owned by other teams. Notably, the ATR mechanism can partially address coverage gaps over time, as new traversals through previously uncovered regions synthesize edges and deposit reinforcement, expanding the graph's effective coverage.

\smallskip\noindent\textbf{Summary.}
Across the 143 native-orchestrator conversations, \sysname achieves interactive latency (median 7.7\,s) and high query success (83\%). Among the 62 scored conversations, 97\% yield actionable output, validating that offline evaluation gains translate to real-world triage workflows.

\smallskip\noindent\textbf{Lessons Learned.}
(1)~\emph{A perfect seed graph is difficult to achieve}: despite extensive prompt refinement, the offline extractor still encounters novel edge cases, especially when onboarding new service teams. Online agentic traversal and ATR evolution help the LLM select effective actions and reasoning paths from an imperfect graph, making careful tuning of the traversal algorithm increasingly important.
(2)~\emph{Graph structure matters more than retrieval volume}: Agentic Summary-RAG retrieves more source incidents yet achieves lower quality, confirming that cross-incident deduplication and explicit causal edges provide a more efficient knowledge representation than per-incident summaries.
(3)~\emph{Graphs serve engineers, not just agents}: beyond powering autonomous agents and MCP-based tool integrations, engineering teams have requested direct access to the underlying graph database for manual review and on-call education. Generating structured documentation from the graph has therefore emerged as a valuable new component, as well as a convenient way to explore the graph.
(4)~\emph{Unifying heterogeneous knowledge sources}: operational traces are only one knowledge modality; existing documentation, runbooks, and design docs contain complementary information. A more general extraction pipeline that ingests diverse sources into the same graph schema is an important direction to further improve graph coverage.

%% file: 6a-discussion.tex

%% file: 8-conclusions.tex
\section{Conclusions}

We presented \sysname, a system that transforms operational traces into self-improving workflow graphs using reinforcement dynamics inspired by Ant Colony Optimization.
Through three pillars---offline construction, online multi-agent traversal, and Adaptive Traversal Reinforcement---the system demonstrates that workflow graphs can continuously learn from autonomous execution, outperforming flat retrieval baselines in mitigation reach, hallucination rate, and diagnostic throughput while requiring 8$\times$ less retrieval context.
More broadly, this action-centric representation of operational knowledge opens up a rich design space for future reinforcement and adaptive learning mechanisms, enabling workflow graphs to evolve continuously through interaction, execution feedback, and long-term system usage patterns.

\smallskip\noindent\textbf{Future Work.}
First, \emph{learning from low-quality trajectories}: applying negative reinforcement to suppress edges along failed trajectories could accelerate convergence.
Second, \emph{exploration vs.\ exploitation}: as graphs grow, reinforcement may over-concentrate on high-traffic subgraphs, necessitating strategies to surface underused paths.
Third, \emph{domain-differentiated reinforcement}: inspired by colony-specific pheromone signatures~\cite{dettorre2010nestmate}, maintaining separate weight channels per problem domain would reduce cross-domain noise to further improve accuracy.

%% file: 9-genai-disclosure.tex
\section{GenAI Usage Disclosure}

ChatGPT and GitHub Copilot were utilized to assist in generating sections of this work, including text, tables, graphs, code, data, and citations.

%% file: sample.bib
@String{Computing = "Computing" }

@String{Computer = "{IEEE} Computer" }

@String{Psychometrika = "Psychometrika" }

@String{Springer = "Springer-Verlag" }

@article{vanderaalst2004workflow,
  title   = {Workflow Mining: Discovering Process Models from Event Logs},
  author  = {van der Aalst, Wil M. P. and Weijters, Ton and Maruster, Laura},
  journal = {IEEE Transactions on Knowledge and Data Engineering},
  volume  = {16},
  number  = {9},
  pages   = {1128--1142},
  year    = {2004},
  doi     = {10.1109/TKDE.2004.47}
}

@book{vanderaalst2012processmining,
  title     = {Process Mining: Discovery, Conformance and Enhancement of Business Processes},
  author    = {van der Aalst, Wil M. P.},
  publisher = {Springer},
  year      = {2011},
  doi       = {10.1007/978-3-642-19345-3}
}

@article{pan2024unifying,
  title   = {Unifying Large Language Models and Knowledge Graphs: A Roadmap},
  author  = {Pan, Shirui and Luo, Linhao and Wang, Yufei and Chen, Chen and Wang, Jiapu and Wu, Xindong},
  journal = {IEEE Transactions on Knowledge and Data Engineering},
  volume  = {36},
  number  = {7},
  pages   = {3580--3601},
  year    = {2024},
  doi     = {10.1109/TKDE.2024.3352100}
}

@inproceedings{kgsft2025,
  title     = {Knowledge Graph Finetuning Enhances Knowledge Manipulation in Large Language Models},
  author    = {Xia, Xinyi and Zhu, Yijun and Guo, Jianan and others},
  booktitle = {Proceedings of the International Conference on Learning Representations (ICLR)},
  year      = {2025},
  url       = {https://openreview.net/forum?id=oMFOKjwaRS}
}

@article{han2025graphrag,
  title   = {Retrieval-Augmented Generation with Graphs ({GraphRAG})},
  author  = {Han, Haoyu and Wang, Yu and Shomer, Harry and Guo, Kai and Ding, Jiayuan and Lei, Yongjia and Halappanavar, Mahantesh and Rossi, Ryan A. and Mukherjee, Subhabrata and others},
  journal = {arXiv preprint arXiv:2501.00309},
  year    = {2025}
}

@misc{edge2024graphrag,
      title={From Local to Global: A Graph RAG Approach to Query-Focused Summarization}, 
      author={Darren Edge and Ha Trinh and Newman Cheng and Joshua Bradley and Alex Chao and Apurva Mody and Steven Truitt and Dasha Metropolitansky and Robert Osazuwa Ness and Jonathan Larson},
      year={2025},
      eprint={2404.16130},
      archivePrefix={arXiv},
      primaryClass={cs.CL},
      url={https://arxiv.org/abs/2404.16130}, 
}

@inproceedings{ograg2024,
  title     = {{OG-RAG}: Ontology-Grounded Retrieval-Augmented Generation for Large Language Models},
  author    = {Ravichandran, Kartik and Kumar, Namrata Gurumurthy and Mishra, Prateek and Agrawal, Rahul},
  booktitle = {Proceedings of the Conference on Empirical Methods in Natural Language Processing (EMNLP)},
  year      = {2025},
  url       = {https://arxiv.org/abs/2412.15235},
  doi       = {10.18653/v1/2025.emnlp-main.1674}
}

@inproceedings{kovriguina2023sparql,
  title     = {{LLM}-based {SPARQL} Query Generation from Natural Language over Federated Knowledge Graphs},
  author    = {Kovriguina, Liubov and Toma, Irina and others},
  booktitle = {Proceedings of the International Semantic Web Conference (ISWC)},
  year      = {2024},
  url       = {https://arxiv.org/abs/2410.06062}
}

@article{delong2024neurosymbolic,
  title   = {Neurosymbolic {AI} for Reasoning Over Knowledge Graphs: A Survey},
  author  = {DeLong, Lara N. and Mir, Ramon Fernandez and Fleuriot, Jacques D.},
  journal = {IEEE Transactions on Neural Networks and Learning Systems},
  year    = {2024},
  doi     = {10.1109/TNNLS.2024.3420218}
}

@article{zhou2024dbot,
  title     = {{D-Bot}: Database Diagnosis System using Large Language Models},
  author    = {Zhou, Xuanhe and Li, Guoliang and Liu, Zhaoyan and others},
  journal   = {Proceedings of the VLDB Endowment},
  volume    = {17},
  number    = {10},
  pages     = {2514--2527},
  year      = {2024},
  doi       = {10.14778/3675034.3675043}
}

@misc{wang2026cloudopsbench,
      title={Cloud-OpsBench: A Reproducible Benchmark for Agentic Root Cause Analysis in Cloud Systems}, 
      author={Yilun Wang and Guangba Yu and Haiyu Huang and Zirui Wang and Yujie Huang and Pengfei Chen and Michael R. Lyu},
      year={2026},
      eprint={2603.00468},
      archivePrefix={arXiv},
      primaryClass={cs.SE},
      url={https://arxiv.org/abs/2603.00468}, 
}

@misc{ni2026trace2skill,
      title={Trace2Skill: Distill Trajectory-Local Lessons into Transferable Agent Skills}, 
      author={Jingwei Ni and Yihao Liu and Xinpeng Liu and Yutao Sun and Mengyu Zhou and Pengyu Cheng and Dexin Wang and Erchao Zhao and Xiaoxi Jiang and Guanjun Jiang},
      year={2026},
      eprint={2603.25158},
      archivePrefix={arXiv},
      primaryClass={cs.AI},
      url={https://arxiv.org/abs/2603.25158}, 
}

@inproceedings{yao2023react,
  title     = {{ReAct}: Synergizing Reasoning and Acting in Language Models},
  author    = {Yao, Shunyu and Zhao, Jeffrey and Yu, Dian and Du, Nan and Shafran, Izhak and Narasimhan, Karthik and Cao, Yuan},
  booktitle = {Proceedings of the International Conference on Learning Representations (ICLR)},
  year      = {2023}
}

@inproceedings{schick2023toolformer,
  title     = {Toolformer: Language Models Can Teach Themselves to Use Tools},
  author    = {Schick, Timo and Dwivedi-Yu, Jane and Dess{\`\i}, Roberto and Raileanu, Roberta and Lomeli, Maria and Hambro, Eric and Zettlemoyer, Luke and Cancedda, Nicola and Scialom, Thomas},
  booktitle = {Advances in Neural Information Processing Systems (NeurIPS)},
  year      = {2023}
}

@misc{openai2023functioncalling,
  title        = {Function Calling},
  author       = {{OpenAI}},
  year         = {2023},
  howpublished = {\url{https://platform.openai.com/docs/guides/function-calling}},
  note         = {Accessed: 2025-05-15}
}

@article{johnson1967hierarchical,
  title   = {Hierarchical Clustering Schemes},
  author  = {Johnson, Stephen C.},
  journal = {Psychometrika},
  volume  = {32},
  number  = {3},
  pages   = {241--254},
  year    = {1967},
  month   = sep,
  doi     = {10.1007/BF02289588}
}

@misc{mcp,
  title        = {Model Context Protocol (MCP)},
  author       = {{Anthropic}},
  year         = {2024},
  month        = nov,
  howpublished = {\url{https://modelcontextprotocol.io/}},
  note         = {Open standard for connecting AI applications to external systems; enables structured, secure interaction between LLMs and data sources/tools via a universal protocol.},
}

@inproceedings{triangle2025,
  title        = {TRIANGLE: A Benchmark and Framework for Automated Incident Triage in Large-Scale Cloud Systems},
  author       = {Xue, Yuqing and Wang, Zhuoran and Sun, Wei and Meng, Fanxin and Zhang, Wenchao and Li, Zhenyu and others},
  booktitle    = {Proceedings of the ACM Joint European Software Engineering Conference and Symposium on the Foundations of Software Engineering (ESEC/FSE)},
  year         = {2025},
  note         = {\url{https://netman.aiops.org/wp-content/uploads/2025/10/TRIANGLE_FSE25.pdf}}
}

@inproceedings{lewis2020rag,
  title        = {Retrieval-Augmented Generation for Knowledge-Intensive NLP Tasks},
  author       = {Lewis, Patrick and Perez, Ethan and Piktus, Aleksandra and Petroni, Fabio and Karpukhin, Vladimir and Goyal, Naman and Küttler, Heinrich and Lewis, Mike and Yih, Wen-tau and Rocktäschel, Tim and others},
  booktitle    = {Advances in Neural Information Processing Systems (NeurIPS)},
  year         = {2020}
}

@inproceedings{zhu2025decolifecyclemanagemententerprisegrade,
author = {Zhu, Yiwen and Demarne, Mathieu and Deng, Kai and Wang, Wenjing and Sahoo, Nutan and Lerner, Hannah and Bhavan, Anjali and Vermareddy, Divya and Lu, Yunlei and Bararia, Swati and Zhang, William and Li, Xia and Lin, Katherine and Cilimdzic, Miso and Krishnan, Subru},
title = {ENCO: Deploying Production-Scale Engineering Copilots},
year = {2026},
isbn = {9798400722585},
publisher = {Association for Computing Machinery},
address = {New York, NY, USA},
url = {https://doi.org/10.1145/3770854.3783949},
doi = {10.1145/3770854.3783949},
booktitle = {Proceedings of the 32nd ACM SIGKDD Conference on Knowledge Discovery and Data Mining V.1},
pages = {2606–2616},
numpages = {11},
location = {Republic of Korea},
series = {KDD '26}
}

@inproceedings{flair2025,
  title        = {FLAIR: Feedback Learning for Adaptive Information Retrieval},
  author       = {Zhang, William and Zhu, Yiwen and Lu, Yunlei and Demarne, Mathieu and Wang, Wenjing and Deng, Kai and Sahoo, Nutan and Lin, Katherine and Cilimdzic, Miso and Krishnan, Subru},
  booktitle    = {Proceedings of the 34th ACM International Conference on Information and Knowledge Management (CIKM '25)},
  year         = {2025},
  address      = {Seoul, Republic of Korea},
  publisher    = {Association for Computing Machinery},
  doi          = {10.1145/3746252.3761553},
  isbn         = {979-8-4007-2040-6/2025/11},
  url          = {https://doi.org/10.1145/3746252.3761553}
}

@misc{openai2023gpt4,
  title        = {GPT-4 Technical Report},
  author       = {{OpenAI}},
  year         = {2023},
  howpublished = {\url{https://cdn.openai.com/papers/gpt-4.pdf}}
}

@inproceedings{brown2020language,
  title        = {Language Models are Few-Shot Learners},
  author       = {Brown, Tom and Mann, Benjamin and Ryder, Nick and Subbiah, Melanie and Kaplan, Jared D and Dhariwal, Prafulla and Neelakantan, Arvind and Shyam, Pranav and Sastry, Girish and Askell, Amanda and others},
  booktitle    = {Advances in Neural Information Processing Systems (NeurIPS)},
  year         = {2020}
}

@inproceedings{shahkgqa,
    title = "Improving {LLM}-based {KGQA} for multi-hop Question Answering with implicit reasoning in few-shot examples",
    author = "Shah, Mili  and
      Cahoon, Joyce  and
      Milletari, Mirco  and
      Tian, Jing  and
      Psallidas, Fotis  and
      Mueller, Andreas  and
      Litombe, Nick",
    booktitle = "Proceedings of the 1st Workshop on Knowledge Graphs and Large Language Models (KaLLM 2024)",
    month = aug,
    year = "2024",
    address = "Bangkok, Thailand",
    publisher = "Association for Computational Linguistics",
    url = "https://aclanthology.org/2024.kallm-1.13/",
    doi = "10.18653/v1/2024.kallm-1.13",
    pages = "125--135"
}

@misc{githubcopilotcli,
  title        = {GitHub Copilot CLI},
  author       = {{GitHub}},
  year         = {2026},
  howpublished = {\url{https://github.blog/changelog/2026-02-25-github-copilot-cli-is-now-generally-available/}},
  note         = {MCP-compatible agentic coding assistant for the command line.}
}

@misc{claudecode,
  title        = {Claude Code},
  author       = {{Anthropic}},
  year         = {2025},
  howpublished = {\url{https://docs.anthropic.com/en/docs/claude-code}},
  note         = {Agentic coding tool with built-in MCP client support.}
}

@article{wang2024survey,
  title={A survey on large language model based autonomous agents},
  author={Wang, Lei and Ma, Chen and Feng, Xueyang and Zhang, Zeyu and Yang, Hao and Zhang, Jingsen and Chen, Zhiyuan and Tang, Jiakai and Chen, Xu and Lin, Yankai and others},
  journal={Frontiers of Computer Science},
  volume={18},
  number={6},
  pages={186345},
  year={2024},
  publisher={Springer},
  url={https://doi.org/10.1007/s11704-024-40231-1}
}

@article{gini1921measurement,
  title={Measurement of inequality of incomes},
  author={Gini, Corrado},
  journal={The economic journal},
  volume={31},
  number={121},
  pages={124--125},
  year={1921},
  publisher={Oxford University Press Oxford, UK}
}

@ArtifactSoftware{R,
    title = {R: A Language and Environment for Statistical Computing},
    author = {{R Core Team}},
    organization = {R Foundation for Statistical Computing},
    address = {Vienna, Austria},
    year = {2019},
    url = {https://www.R-project.org/},
}

@article{singh2025agenticrag,
  title={Agentic Retrieval-Augmented Generation: A Survey on Agentic {RAG}},
  author={Aditi Singh and Abul Ehtesham and Saket Kumar and Tala Talaei Khoei},
  journal={arXiv preprint arXiv:2501.09136},
  year={2025}
}

@inproceedings{asai2024selfrag,
  title={Self-{RAG}: Learning to Retrieve, Generate, and Critique through Self-Reflection},
  author={Akari Asai and Zeqiu Wu and Yizhong Wang and Avirup Sil and Hannaneh Hajishirzi},
  booktitle={Proceedings of the International Conference on Learning Representations (ICLR)},
  year={2024}
}

@inproceedings{singh2024panda,
  title     = {Panda: Performance Debugging for Databases using {LLM} Agents},
  author    = {Vikramank Singh and Kapil Eknath Vaidya and Vinayshekhar Bannihatti Kumar and Sopan Khosla and Murali Narayanaswamy and Rashmi Gangadharaiah and Tim Kraska},
  booktitle = {Proceedings of the Conference on Innovative Data Systems Research (CIDR)},
  year      = {2024}
}

@inproceedings{chen2024rcacopilot,
  title     = {Automatic Root Cause Analysis via Large Language Models for Cloud Incidents},
  author    = {Yinfang Chen and Huaibing Xie and Minghua Ma and Yu Kang and Xin Gao and Liu Shi and Yunjie Cao and Xuedong Gao and Hao Fan and Ming Wen and Jun Zeng and Supriyo Ghosh and Xuchao Zhang and Chaoyun Zhang and Qingwei Lin and Saravan Rajmohan and Dongmei Zhang and Tianyin Xu},
  booktitle = {Proceedings of the 19th European Conference on Computer Systems (EuroSys)},
  pages     = {674--688},
  year      = {2024},
  doi       = {10.1145/3627703.3629553}
}

@article{dorigo1996aco,
  title   = {Ant system: optimization by a colony of cooperating agents},
  author  = {Marco Dorigo and Vittorio Maniezzo and Alberto Colorni},
  journal = {IEEE Transactions on Systems, Man, and Cybernetics, Part B (Cybernetics)},
  volume  = {26},
  number  = {1},
  pages   = {29--41},
  year    = {1996},
  doi     = {10.1109/3477.484436}
}

@book{dorigo2004aco,
  title     = {Ant Colony Optimization},
  author    = {Marco Dorigo and Thomas St{\"u}tzle},
  publisher = {MIT Press},
  year      = {2004}
}

@article{stutzle2000maxmin,
  title   = {{MAX-MIN} Ant System},
  author  = {Thomas St{\"u}tzle and Holger H. Hoos},
  journal = {Future Generation Computer Systems},
  volume  = {16},
  number  = {8},
  pages   = {889--914},
  year    = {2000},
  doi     = {10.1016/S0167-739X(00)00043-1}
}

@inproceedings{dorigo1997acs,
  title     = {Ant colony system: a cooperative learning approach to the traveling salesman problem},
  author    = {Marco Dorigo and Luca Maria Gambardella},
  booktitle = {IEEE Transactions on Evolutionary Computation},
  volume    = {1},
  number    = {1},
  pages     = {53--66},
  year      = {1997},
  doi       = {10.1109/4235.585892}
}

@article{dicaro1998antnet,
  title   = {{AntNet}: distributed stigmergetic control for communications networks},
  author  = {Gianni {Di Caro} and Marco Dorigo},
  journal = {Journal of Artificial Intelligence Research},
  volume  = {9},
  pages   = {317--365},
  year    = {1998}
}

@inproceedings{merkle2002aco_scheduling,
  title     = {Ant colony optimization for resource-constrained project scheduling},
  author    = {Daniel Merkle and Martin Middendorf and Hartmut Schmeck},
  booktitle = {IEEE Transactions on Evolutionary Computation},
  volume    = {6},
  number    = {4},
  pages     = {333--346},
  year      = {2002}
}

@article{tabakhi2014aco_feature,
  title   = {An unsupervised feature selection algorithm based on ant colony optimization},
  author  = {Sina Tabakhi and Parham Moradi and Fardin Akhlaghian},
  journal = {Engineering Applications of Artificial Intelligence},
  volume  = {32},
  pages   = {112--123},
  year    = {2014}
}

@inproceedings{bordes2013transe,
  title     = {Translating Embeddings for Modeling Multi-relational Data},
  author    = {Antoine Bordes and Nicolas Usunier and Alberto Garcia-Duran and Jason Weston and Oksana Yakhnenko},
  booktitle = {Advances in Neural Information Processing Systems (NeurIPS)},
  year      = {2013}
}

@inproceedings{dettmers2018conve,
  title     = {Convolutional 2D Knowledge Graph Embeddings},
  author    = {Tim Dettmers and Pasquale Minervini and Pontus Stenetorp and Sebastian Riedel},
  booktitle = {Proceedings of the AAAI Conference on Artificial Intelligence},
  year      = {2018}
}

@inproceedings{wang2014transr,
  title     = {Knowledge Graph Embedding by Translating on Hyperplanes},
  author    = {Zhen Wang and Jianwen Zhang and Jianlin Feng and Zheng Chen},
  booktitle = {Proceedings of the AAAI Conference on Artificial Intelligence},
  year      = {2014}
}

@incollection{dettorre2010nestmate,
  title     = {Nestmate Recognition},
  author    = {d'Ettorre, Patrizia and Lenoir, Alain},
  booktitle = {Ant Ecology},
  editor    = {Lach, Lori and Parr, Catherine L. and Abbott, Kirsti L.},
  publisher = {Oxford University Press},
  year      = {2010},
  pages     = {194--209}
}

@inproceedings{ma2024spreadsheetbench,
  author    = {Ma, Zeyao and Zhang, Bohan and Zhang, Jing and Yu, Jifan and Zhang, Xiaokang and Zhang, Xiaohan and Luo, Sijia and Wang, Xi and Tang, Jie},
  booktitle = {Advances in Neural Information Processing Systems},
  doi       = {10.52202/079017-3007},
  pages     = {94871--94908},
  publisher = {Curran Associates, Inc.},
  title     = {{SpreadsheetBench}: Towards Challenging Real World Spreadsheet Manipulation},
  volume    = {37},
  year      = {2024}
}
